\documentclass[a4paper,fleqn]{cas-dc}
\usepackage[authoryear]{natbib}
\usepackage{lineno} 
\usepackage{multirow}
\usepackage{cuted}
\usepackage{enumitem} 
\usepackage{booktabs}
\usepackage{tabularx}

\usepackage{graphicx}
\usepackage{xcolor}
\usepackage{makecell}
\usepackage{multirow}
\usepackage{booktabs}
\usepackage{array}
\usepackage{tabularx}
\newcolumntype{Y}{>{\centering\arraybackslash}X}

\AtBeginDocument{
  \setlength{\abovedisplayskip}{4pt}
  \setlength{\belowdisplayskip}{4pt}
  \setlength{\abovedisplayshortskip}{2pt}
  \setlength{\belowdisplayshortskip}{2pt}
}

\usepackage{etoolbox}
\AfterEndPreamble{
  \makeatletter
  \let\oldthebibliography\thebibliography
  \let\endoldthebibliography\endthebibliography
  
  \renewenvironment{thebibliography}[1]
    {\oldthebibliography{#1}%
     \small%
     \setlength{\itemsep}{2pt}%
     \setlength{\topsep}{6pt}%
    }
    {\endoldthebibliography}
  \makeatother
}

\usepackage{array}
\newcommand{\viscell}[2]{\shortstack[c]{\strut #1\\[-1pt]\scriptsize(#2\,)\strut}}

\def\tsc#1{\csdef{#1}{\textsc{\lowercase{#1}}\xspace}}
\tsc{WGM}
\tsc{QE}

\begin{document}
\let\WriteBookmarks\relax
\def\floatpagepagefraction{1}
\def\textpagefraction{.001}

\shorttitle{}    

\shortauthors{}  

\title [mode = title]{Bridging the Gap Between Image Restoration and Navigational Safety in Hazy Conditions: A New Visibility Estimation Metric for Maritime Surveillance}  



%

\author[1,2,3]{Wentao Feng}

\ead{fengwentao@whut.edu.cn}
\author[1,2,3]{Guobei Peng}

\ead{pengguobei@whut.edu.cn}

\author[4]{Wengang Mao}

\ead{wengang.mao@chalmers.se}

\author[1,2,3]{Ryan Wen Liu\corref{cor1}}

\ead{wenliu@whut.edu.cn}

\cortext[cor1]{Corresponding author.} 

\address[1]{Sanya Science and Education Innovation Park, Wuhan University of Technology, Sanya 572000, China}
\address[2]{School of Navigation, Wuhan University of Technology, Wuhan 430063, China}
\address[3]{State Key Laboratory of Maritime Technology and Safety, Wuhan University of Technology, Wuhan 430063, China}
\address[4]{Department of Mechanics and Maritime Sciences, Chalmers University of Technology, Göteborg 41296, Sweden}














\begin{abstract}
Visibility distance is a critical factor for maritime navigational safety, as it determines the effective observation range of shipborne and shore-based monitoring systems. Under hazy conditions, the degradation of visual information significantly reduces the observable distance, leading to increased navigational risk and economic loss. Although numerous image dehazing methods have been developed to enhance visual quality, existing image quality assessment (IQA) metrics (e.g., PSNR, SSIM, FSIM, FADE, NIQE, etc.) fail to provide a physically interpretable link between image restoration quality and actual navigational safety thresholds, even though the dehazing performance could be undoubtedly obtained in maritime practice. This limitation arises because these objective metrics do not reliably reflect absolute visibility levels or effective observation range. Motivated by the strong correlation between navigational safety and visible distance,  this work proposes a visibility-oriented evaluation framework that links image enhancement performance with practical visibility estimation. In particular, to alleviate the issue of lacking hazy and clear image pairs for dehazing networks, a Maritime Simulated Visibility Dataset (MSVD) is constructed using the Unity3D physics engine to simulate maritime traffic scenes under graded visibility conditions. The dataset provides paired hazy–clear images together with precise visibility annotations, enabling quantitative analysis of visibility restoration. Besides, a new dehazing visibility evaluation metric is proposed by leveraging object detection performance as an intermediate indicator. By establishing the relationship between visibility distance and detection accuracy, the proposed metric translates improvements in image restoration into measurable visibility gains. Six different dehazing methods are then utilized to perform visibility restoration. The dehazing performance and visible distance estimation are quantitatively compared using traditional image quality assessment metrics and our visibility evaluation metric. Experimental results in different imaging conditions demonstrate that (1) our dataset MSVD provides a reliable benchmark for evaluating dehazing performance across graded visibility levels; (2) our visibility evaluation metric contributes to highly-reliable estimation of visible distance, thereby supporting a balanced trade-off between navigational safety and operational efficiency. \nocite{*}
\end{abstract}


\begin{highlights}
\item Propose a visibility-oriented evaluation metric for maritime dehazing.

\item Construct a physics-consistent Maritime Simulated Visibility Dataset (MSVD).

\item Analyze the limitations of existing image quality metrics in maritime visibility evaluation.

\item Demonstrate the robustness of our metric across multiple dehazing and detection models.
\end{highlights}



\maketitle


\section{Introduction}

In maritime transportation, maintaining effective visual observation is a fundamental requirement for safe navigation and collision avoidance \citep{WANG2023114329,WU2025120616}. Both shipborne cameras and shore-based monitoring systems are widely deployed to support traffic surveillance, vessel tracking, and situational awareness in coastal waters \citep{liu2021enhanced}. The operational capability of these systems is largely determined by the effective visible distance that can be achieved under varying atmospheric conditions. 
\begin{figure}
\centering
\includegraphics[width=0.88\columnwidth]{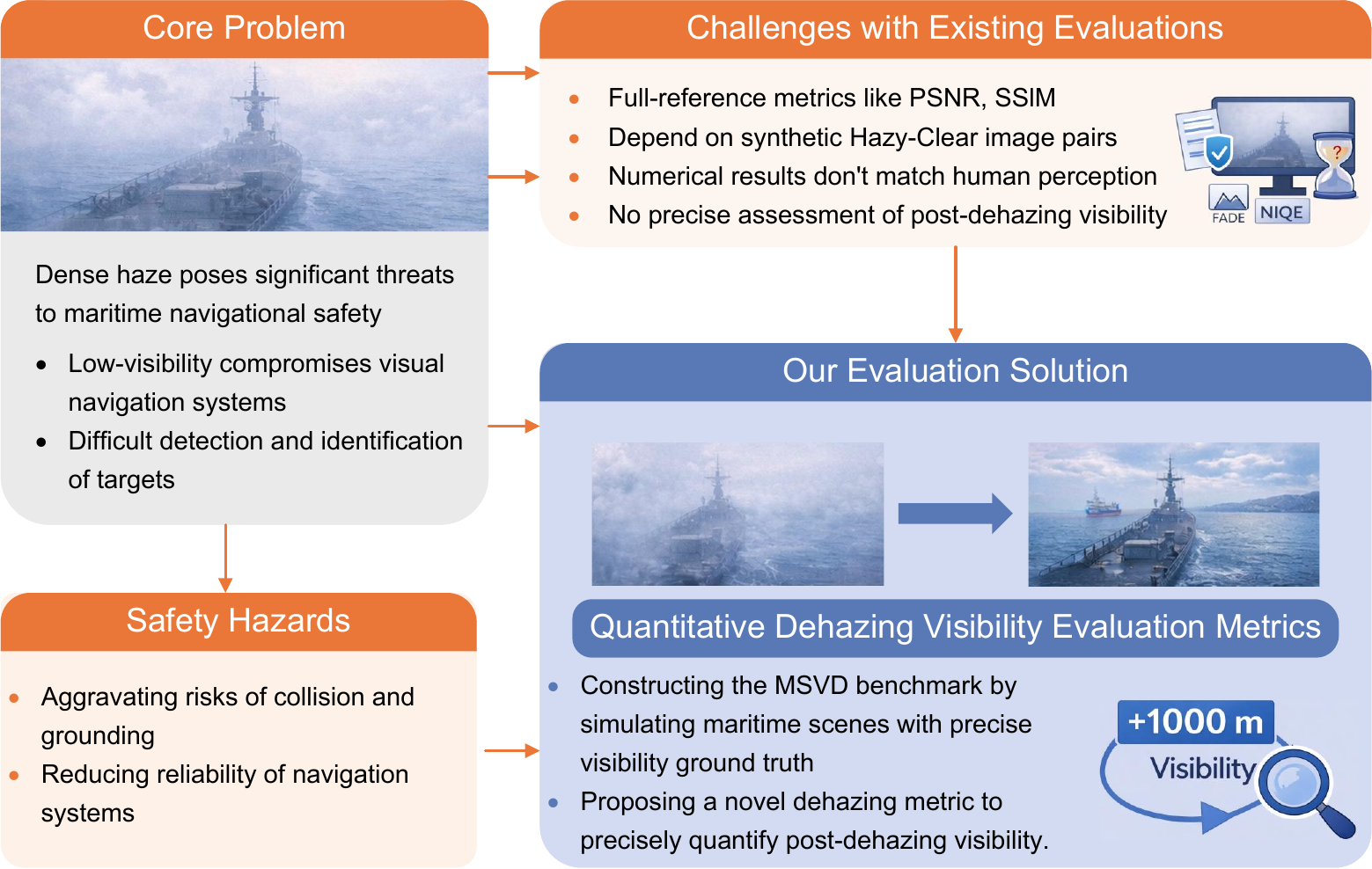}
\caption{The framework of the maritime dehazing visibility distance evaluation system. It delineates the transition from identifying maritime navigational risks and the limitations of current metrics to the proposed solution.}
\label{fig:1}
\end{figure}
However, in real maritime environments, haze frequently degrades image contrast and reduces the observable range of optical sensors, making distant targets difficult to be detected and recognized \citep{he2025complexity}. When visibility deteriorates, the performance of monitoring systems may decline significantly, potentially increasing navigational risks and reducing traffic management efficiency. 

\begin{table}[t]
\centering
\renewcommand{\arraystretch}{1.1}
\caption{Visibility distance classification and corresponding navigation restrictions under hazy conditions.}
\label{tab:visibility}

\footnotesize

\begin{tabular}{>{\centering\arraybackslash}m{0.23\columnwidth}
                >{\centering\arraybackslash}m{0.25\columnwidth}
                >{\raggedright\arraybackslash}m{0.38\columnwidth}}

\toprule
\textbf{Haze Density} & \textbf{Visibility Distance} & \textbf{Navigation Regulation} \\
\midrule
\rule{0pt}{3.3ex} Very Dense Haze &  < 50 m & Navigation Suspended \\
\rule{0pt}{3.3ex} Dense Haze &  < 200 m & Near Suspension of Navigation \\
\rule{0pt}{3.3ex} Moderate Haze &  < 500 m & Strict Navigation Restrictions \\
\rule{0pt}{3.3ex} Light Haze &  < 1000 m & Restricted Navigation \\
\rule{0pt}{3.3ex} Thin Haze &  < 2000 m & Minor Restrictions \\
\rule{0pt}{3.3ex} Poor Visibility &  < 4000 m & Normal Navigation \\
\bottomrule

\end{tabular}
\end{table}

From a regulatory perspective, maritime authorities rely on prescribed visibility thresholds when determining operational measures that ensure navigational safety \citep{liu2024real}. Higher safety requirements typically demand longer operational visibility distances, which may result in stricter traffic control strategies and reduced traffic efficiency \citep{lu2026graph}. Consequently, accurately measuring visibility recovery is not only a technical issue but also an important factor in balancing navigational safety and operational efficiency. For instance, as categorized in Table \ref{tab:visibility}, a precise leap in estimated visibility distance, such as from a restricted threshold (e.g., < 1000m) to a maneuverable range (e.g., > 2000m) through dehazing, can directly empower maritime authorities to transition from restricted navigation to normal navigation. Without a quantifiable visibility metric, the performance gain of image enhancement remains a black box, making it impossible for regulators to safely relax restrictions based on enhanced visual data. These operational requirements indicate that maritime image enhancement techniques should be evaluated according to their ability to restore effective observation distance and support reliable object detection, rather than solely by improvements in visual quality.

To alleviate the degradation of visual information caused by haze, a large number of image dehazing methods have been proposed in recent years \citep{huang2025dual}. Early approaches mainly relied on physical atmospheric scattering models and handcrafted priors, such as the Dark Channel Prior and contrast enhancement strategies. More recently, deep learning-based methods have achieved substantial progress by learning complex haze removal mappings from large-scale datasets. These techniques are capable of significantly improving visual clarity and recovering structural details in degraded images. 

Despite these technical advances, the practical value of dehazing methods in maritime applications remains difficult to quantify. A fundamental challenge lies in how the performance of these algorithms should be evaluated. In most existing studies, the effectiveness of dehazing methods is assessed using traditional image quality assessment metrics. These metrics can be broadly categorized into Full-Reference (FR) and No-Reference (NR) metrics. FR metrics, such as PSNR \citep{5596999} and SSIM \citep{wang2004image}, require haze-free ground truth images for comparison, whereas NR metrics, including FADE \citep{choi2015referenceless} and NIQE \citep{mittal2012making}, attempt to estimate perceptual image quality without reference data. Although these metrics assess image fidelity or perceptual quality, they cannot reflect the effective observation distance within the scene. Consequently, improvements in these scores may not reflect genuine enhancements in visibility distance. This mismatch becomes particularly critical in maritime environments. For maritime authorities and Vessel Traffic Services (VTS), visibility distance serves as the core parameter for determining navigation control strategies, directly influencing decisions such as traffic regulation levels and temporary channel closures. Operational decisions are often based on predefined visibility thresholds specified in maritime safety regulations. Therefore, from a practical perspective, the effectiveness of image enhancement methods should ideally be measured according to their ability to restore observable distance and support reliable object detection.

Metrics that only evaluate visual quality cannot fully capture this operational requirement. Despite the existence of dedicated visibility estimation methods, they predominantly rely on the DCP prior that frequently fails in maritime environments. Conversely, deep learning-based estimation methods are hindered by a severe lack of paired maritime training datasets. Acquiring real-world paired hazy and haze-free images with accurate physical annotations in these settings is notoriously difficult. While synthetic datasets are widely used, they primarily focus on urban traffic scenarios and lack precise visibility distance labels. Consequently, existing datasets remain insufficient for studying the relationship between image restoration performance and visibility distance in maritime navigation.

Based on the aforementioned challenges, two core problems must be addressed to bridge the gap between technical efficacy and practical utility:

{\bf{1) How to acquire paired maritime dehazing datasets with precise physical visibility distance labels?}} In real-world maritime environments, obtaining corresponding haze-free images as ground truth is extremely difficult. Although synthetic datasets help address this problem, they usually lack reliable physical annotations, particularly accurate visibility distance or depth information. In conclusion, they are not well suited for the quantitative evaluation of dehazing methods in safety-critical maritime applications.

{\bf{2) How to quantitatively evaluate visibility distance improvement after the dehazing process?}} Neither FR nor NR metrics can directly address the operationally critical question of visibility gain. Specifically, they fail to determine the magnitude of actual visibility distance improvement. Existing numerical results often deviate from human perceptual judgment and lack the ability to map evaluation scores to physically interpretable quantities, thereby failing to establish a direct linkage between dehazing efficacy and maritime regulatory decision-making requirements.

Addressing these challenges requires a unified framework that integrates physical-based datasets with visibility evaluation metrics. On the one hand, without datasets containing accurately calibrated visibility distance annotations, it is difficult to validate whether a proposed visibility metric truly reflects real observation distance. On the other hand, without a dedicated evaluation metric, even high-fidelity datasets cannot be used to quantitatively measure how much visibility distance is improved after dehazing. Therefore, both physically reliable data and appropriate evaluation metrics are necessary to establish a quantitative and operationally meaningful evaluation framework for maritime visibility enhancement. Based on this motivation, this study makes the following contributions:

·A Maritime Simulated Visibility Dataset (MSVD) is constructed using the Unity3D physics engine to generate maritime navigation scenes under graded visibility conditions, providing paired hazy and clear images together with precise visibility distance annotations.

·A new visibility distance evaluation metric is proposed to quantify the improvement in effective visibility distance after image dehazing by linking image restoration performance with object detection index.

·Extensive experiments are conducted to analyze the relationship between dehazing performance and  visibility distance, demonstrating the effectiveness and robustness of the proposed evaluation metric.

\section{Related work}
In maritime navigation, restoring atmospheric visibility is essential for the reliable operation of intelligent perception systems. Under dense haze conditions, limited visibility distance significantly reduces the effective surveillance range of both shipborne and shore-based platforms. This reduction weakens target recognition and increases uncertainty in traffic situational awareness. Such degradation in perception capability poses a direct threat to navigational safety, since maritime authorities and Vessel Traffic Services rely on visibility distance levels when determining traffic control and navigation measures. However, existing dehazing datasets lack precise visibility distance annotations, which prevents quantitative visibility distance evaluation. While traditional image enhancement methods emphasize visual quality improvement, the fundamental requirement in maritime safety is the restoration of effective observation distance under variable atmospheric conditions. Moreover, current research remains limited in its ability to quantify the actual visibility distance gains achieved through dehazing, which is critical for maintaining reliable situational awareness in complex maritime environments. This section first delineates the limitations of existing dehazing datasets, followed by a review of the algorithmic evolution from physical priors to deep learning architectures. Finally, current image quality assessment (IQA) metrics and visibility distance estimation methods are scrutinized to identify key research priorities in maritime image dehazing.

\subsection{Dehazing Dataset}

The rapid advancement of dehazing methods is largely attributable to the support of large-scale, high-quality datasets. In current research, synthetic datasets facilitate the automated generation of hazy-clear pairs through depth or transmission maps, ensuring perfectly aligned images for model training.

Synthetic dehazing datasets play a crucial role in advancing learning-based methods. Prominent among them is RESIDE-SOTS (Synthetic Objective Testing Set) \citep{li2018benchmarking}, widely recognized as a representative benchmark. It provides diverse indoor and outdoor synthetic scenes with varying haze concentrations and depth distributions, and has significantly promoted the development of deep learning-based dehazing approaches. To further obtain high-definition hazy images with higher fidelity, the Haze4k dataset \citep{liu2021synthetic} offers synthetic images with 4K resolution, providing a more substantial benchmark for evaluating contemporary dehazing architectures. In the context of urban driving, the FRIDA \citep{tarel2010improved} and FRIDA2 \citep{tarel2012vision} datasets were introduced for contrast restoration, containing 90 and 330 synthetic images, respectively. However, their relatively small scale limits their suitability for modern data-driven models. The Cityscapes \citep{cordts2016cityscapes} dataset offers high-resolution street scenes that include adverse weather conditions such as haze and rain. Nevertheless, its application remains largely confined to urban road environments. 

\begin{table}[t]
    \centering
    \renewcommand{\arraystretch}{1.2} 
    \caption{Comparison of various dehazing datasets and the proposed MSVD.} 
    \label{tab:datasets_comparison}
    \small  
    \begin{tabular}{lcccc}
        \toprule
        \textbf{Dataset} & \textbf{Scene} & \textbf{Anchor} & \textbf{Number} & \textbf{\makecell{Visibility \\ Distance}} \\
        \midrule
        \makecell[l]{RESIDE-SOTS \\ \citep{li2018benchmarking}}  & Urban    & Yes & 13,990 & No \\
        \makecell[l]{Cityscapes \\ \citep{cordts2016cityscapes}} & Urban    & Yes & 25,000 & No \\
        \makecell[l]{FRIDA \\ \citep{tarel2012vision}}           & Urban    & No  & 90     & No \\
        \makecell[l]{FRIDA2 \\ \citep{tarel2010improved}}        & Urban    & No  & 330    & No \\
        \makecell[l]{O-HAZE \\ \citep{ancuti2018haze}}           & Outdoor  & No  & 45     & No \\
        \makecell[l]{NH-HAZE \\ \citep{ancuti2020nh}}            & Outdoor  & No  & 55     & No \\
        \makecell[l]{Kede \\ \citep{ma2015perceptual}}           & Outdoor  & No  & 225    & No \\
        \makecell[l]{REMIDE \\ \citep{He2022SingleMI}}           & Maritime & No  & 2,098  & No \\
        \makecell[l]{Overwater-Haze \\ \citep{xie2025overwater}} & Maritime & No  & 21,000 & No \\        
        \midrule
        \textbf{MSVD (Ours)} & \textbf{Maritime} & \textbf{Yes} & \textbf{12,312} & \textbf{Yes} \\
        \bottomrule
    \end{tabular}
\end{table}


While synthetic data can be generated at low cost, the pursuit of realism has driven a shift toward capturing real hazy scenes under controlled condition. The O-HAZE \citep{ancuti2018haze} dataset utilizes professional haze generators to capture 45 outdoor scenes under consistent lighting, providing authentic pairs that bridge the gap between simulation and reality. This data acquisition approach was further refined during the NTIRE2020 challenge, yielding NH-HAZE \citep{ancuti2020nh}. Unlike the homogeneous haze in O-HAZE, NH-HAZE focuses on nonhomogeneous haze, a particularly challenging real-world condition that features spatially varying density.  Beyond the datasets discussed above, Kede \citep{ma2015perceptual} offers further diversity by categorizing images based on outdoor settings and haze thicknesses. Regrettably, these datasets are still largely confined to urban or land-based scenes, which deviate markedly from maritime imaging conditions.

To address the scarcity of maritime dehazing datasets, various researchers have developed specialized hazy image datasets tailored for maritime environments. \citet{He2022SingleMI} constructed a real-world unpaired maritime image dehazing dataset named REMIDE, which contains 2,047 training images (984 hazy and 1,063 clean) collected from maritime scenes. Building upon this, \citet{xie2025overwater} recently introduced Overwater-Haze, a large-scale supervised benchmark comprising 21,000 synthetic paired images and 500 real-world samples. Overwater-Haze provides more rigorous data support for high-precision maritime perception.

The aforementioned datasets provide realistic visual features, however, they often lack precise, objective quantitative ground truth, particularly accurate visibility distance annotations. A comparison between these datasets and the proposed MSVD is presented in Table \ref{tab:datasets_comparison}. This limitation prevents their use in rigorous quantitative evaluations of dehazing performance from a physical perspective. In specialized domains with stringent safety requirements, such as the maritime environment, there remains a severe scarcity of publicly available datasets annotated with precise physical quantities. This gap highlights the need for new benchmarks that can offer both realistic visual data and the physical parameters required for safety-critical tasks, including navigation, collision avoidance, and maritime traffic management.

\subsection{Image Dehazing Methods}
To ensure navigational safety in maritime environments, effective dehazing methods are required to mitigate image degradation caused by haze \citep{chen2026vision}.

The development of image dehazing has progressed from model-driven priors to data-driven deep learning frameworks. Early methods were largely based on priors derived from the atmospheric scattering model. A representative work is the DCP proposed by \citet{he2010single}, which leverages the observation that haze-free images typically contain pixels with very low intensities in at least one color channel within local patches. Along a similar line, \citet{berman2016non} introduced the Non-Local Color Prior (NCP), assuming that colors in a clear image form compact clusters in RGB space. By exploiting the non-local distribution of colors, NCP estimates scene transmission more robustly and alleviates some of the artifacts commonly observed in local-prior-based methods. Although these prior-based methods established a strong theoretical foundation, their performance is sensitive to scene characteristics. For example, DCP often produces artifacts in bright regions such as the sky or water surfaces. Meanwhile, NCP may lead to color distortions or halo effects in complex scenes.  

With the advancement of deep learning, dehazing research gradually shifted toward data-driven paradigms \citep{liu2026real}. Early learning-based methods focused on estimating intermediate parameters of the atmospheric scattering model. For instance, \citet{ren2016single} proposed the Multi-Scale Convolutional Neural Network (MSCNN) to predict the transmission map, and AOD-Net \citep{li2017aod} further integrated transmission and atmospheric light estimation into a unified optimization framework. Subsequent studies moved beyond explicit parameter estimation and treated dehazing as an end-to-end image restoration or image-to-image translation task. GFN \citep{li2018end} employed a gated fusion strategy to integrate features from multiple pre-processed inputs, whereas EPDN \citep{qu2019enhanced} formulated dehazing within an enhanced pix2pix framework and introduced the Perceptual Index (PI) to better align quantitative evaluation with human visual perception. Further advancements emphasized architectural innovations to better address complex degradation patterns. MSBDN \citep{dong2020multi} adopted a boosting strategy for progressive restoration, while 4K-Dehazing \citep{zheng2021ultra} incorporated multi-guided bilateral learning and affine bilateral grids to preserve high-fidelity edge structures in ultra-high-definition images. To enhance robustness under unknown haze distributions, domain-invariant approaches \citep{shyam2021towards} introduced spatially aware channel attention to promote feature consistency.    Several studies also explored multi-scale feature learning and progressive supervision strategies. For example, MSRL-DehazeNet \citep{yeh2019multi} leverages multi-scale residual learning to decompose and refine image components, while LAP-Net \citep{li2019lap} introduces stage-wise supervision to adaptively address varying haze levels within a single scene.   

With the rise of the Vision Transformer (ViT) and attention-driven models like FFA-Net \citep{qin2020ffa}, which utilizes feature attention to capture complex patterns, deep learning methods have achieved state-of-the-art performance in diverse tasks. Recent studies have further explored diverse strategies to improve dehazing performance under complex atmospheric conditions. For example, advanced multi-scale attention mechanisms, combined with photo-realistic synthetic training datasets, have been proposed to enhance the perception and removal of non-uniform dense haze \citep{zhang2024photo}. To alleviate structural distortions and local detail loss, multi-mapping GAN frameworks integrated with probabilistic fuzzy priors have demonstrated improved generalization ability in real-world scenarios. Furthermore, lightweight architectures, such as wavelet-based physically guided normalization networks, have been developed to balance restoration quality and computational efficiency, making them more suitable for real-time applications \citep{zhang2025exploring,zhang2025wavelet}.

Despite their impressive performance, learning-based dehazing methods still face several practical limitations. Most existing approaches rely heavily on large-scale paired training datasets, whose acquisition is particularly challenging in maritime environments. Consequently, their performance and generalization capability are often constrained by the availability and quality of training data.

\label{}
\subsection{Image Quality Assessment}
In the maritime domain, the evaluation of image dehazing performance is inherently linked to operational safety. Traditional research primarily focused on pixel-level restoration quality, which often overlooks the semantic reliability required for vessel detection and collision avoidance. Currently, image quality assessment methods can be classified into two frameworks according to their reliance on a ground-truth reference: FR and NR metrics. 

FR metrics operate by quantifying the fidelity of a dehazed image relative to an ideal clear reference. Traditional benchmarks such as PSNR assess absolute pixel-level errors, while the SSIM incorporates luminance, contrast, and structural information to align more closely with human visual perception. To capture more nuanced visual attributes, researchers have introduced advanced feature-based indices. For instance, to measure haze density variations, \citet{liu2020image} introduced the Fog-relevant Feature-based Similarity index (FRFSIM), which leverages dark channel and contrast-normalized features. Similarly, \citet{min2018objective} proposed a multifaceted approach that employs hand‑crafted features targeting haze removal, structural preservation, and over‑enhancement prevention. With the advent of deep learning, perceptual metrics have gained prominence. \citet{zhang2018unreasonable} demonstrated that deep features, such as those used in the Learned Perceptual Image Patch Similarity (LPIPS), substantially outperform classical metrics. Further developments include the Deep Image Structure and Texture Similarity (DISTS) \citep{ding2020image}, which decouples texture and structural correlations, and PieAPP \citep{prashnani2018pieapp}, which uses a pairwise learning framework to predict human perceptual preference. However, the efficacy of all FR metrics is fundamentally contingent on the availability of perfectly aligned ground-truth images—a condition nearly impossible to satisfy in real-world, dynamic maritime environments.

NR metrics serve as a critical alternative for real-world applications when clear references are unavailable. These metrics often rely on the statistical properties of natural images, such as the Blind/Referenceless Image Spatial Quality Evaluator (BRISQUE) \citep{mittal2011blind} and the Natural Image Quality Evaluator (NIQE) \citep{mittal2012making}. Focusing on haze-specific assessment, \citet{choi2015referenceless} introduced the Fog-Aware Density Evaluator (FADE), which predicts the residual haze density in restored images. Recent advancements have sought to integrate more complex learning mechanisms into the NR framework. For example, \citet{guan2022visibility} developed an NR metric that extracts both visibility and distortion-aware features, mapping them to quality scores via support vector regression. Furthermore, \citet{lv2023blind} introduced the Blind Dehazed Image Quality Model (BDQM), which incorporates a specialized patch attention mechanism to mitigate inhomogeneous distortions inherently introduced by dehazing algorithms. To further address the complex perceptual distortions and residual haze in authentic dehazed images, recent metrics have explored complex-valued network architectures. For instance, \citet{guan2023dual} proposed a dual-stream Complex-Valued Convolutional Neural Network (CV-CNN) for authentic dehazed image quality assessment. This method distinctively employs a distortion-sensitive stream operating on RGB images and a haze-aware stream utilizing novel dark channel difference maps. By extracting complex-valued features, it effectively captures both structural distortions and residual haze, achieving excellent alignment with human subjective perception. Despite these advancements in subjective quality evaluation, existing NR metrics share a fundamental limitation: they primarily assess visual naturalness and fail to bridge the gap between abstract numerical scores and actual physical parameters (e.g., visibility distance). This lack of physical interpretability is particularly problematic in safety-critical maritime domains, significantly limiting their operational value for navigational decision-making.

\label{}
\subsection{Visibility Estimation} 
Image quality assessment metrics primarily focus on visual quality of restoring images. However, the core requirement for maritime safety lies in the accurate estimation of physical visibility distance. Currently, visibility estimation methods can be categorized into two main paradigms: physical model-based methods and deep learning-based methods. 

Physical model-based approaches typically estimate visibility by exploiting geometric constraints or atmospheric scattering priors. A common strategy establishes a relationship between image coordinates and physical distances through vanishing point detection. For example, \citet{bronte2009fog} and \citet{negru2013image} detected road vanishing points using edge extraction and Hough transforms, then computed visibility via camera calibration parameters. Other studies refined transmittance estimation by incorporating the DCP; \citet{bae2019coastal} applied DCP to delineate coastline boundaries, while \citet{graves2011using} estimated visibility through local contrast analysis. Additional improvements, such as guided filtering \citep{he2021visibility} and nonlinear least-squares fitting based on calibrated camera models \citep{zou2017visibility}, were introduced to enhance robustness. Nevertheless, these methods are highly dependent on stable scene geometry and auxiliary calibration parameters. In maritime environments, these foundational assumptions often falter: reflective water surfaces undermine the dark channel prior, whereas the absence of structured landmarks—such as road boundaries—renders reliable vanishing point detection infeasible.

To alleviate reliance on explicit geometric modeling, learning-based approaches extract visibility information directly from high-level visual representations. Early studies incorporated handcrafted descriptors combined with machine learning techniques; for example, \citet{chincholkar2019fog} utilized image brightness and variance as auxiliary features for visibility estimation. Subsequently, hybrid frameworks emerged. \citet{outay2021estimating} integrated deep CNN-based feature extraction with Support Vector Machine (SVM) classification, while \citet{qin2021end} proposed TVRNet, an end-to-end regression architecture designed to capture multi-scale haze density characteristics. Although these data-driven models remove the need for explicit geometric constraints, their generalization performance is largely determined by the diversity and quality of the training data annotations.

Although current visibility estimation methods perform well on terrestrial or urban datasets, they face significant limitations when applied to maritime environments. This limitation primarily stems from two distinct challenges. On the one hand, the sea presents unique visual characteristics, such as dynamic sea-sky horizons, strong water surface reflections, and salt-laden atmospheric scattering. These elements cause classical physical assumptions to fail. For instance, highly reflective water surfaces intrinsically lack the "dark pixels" required by the widely used DCP. Similarly, the open water lacks structured landmarks (e.g., road boundaries), rendering reliable vanishing point detection infeasible. On the other hand, while data-driven deep learning models bypass these explicit geometric constraints, they face a severe data bottleneck. Currently, there is a critical scarcity of large-scale, accurately annotated real-world maritime datasets. Existing synthetic datasets typically label haze using abstract scattering coefficients rather than precise, meter-level physical distances. Without these concrete physical ground truths, learning-based methods struggle to establish reliable visibility representations, leading to poor generalization when deployed in real maritime environments. These fundamental limitations underscore the urgent need to establish a dedicated maritime-oriented benchmark and develop quantitative metrics tailored specifically for complex navigational environments.

\label{}

\section{MSVD: The Maritime Dehazing Dataset with Accurate Visibility Distance Generated by Unity3D}
\begin{figure}
    \centering
    \includegraphics[width=0.95\linewidth]{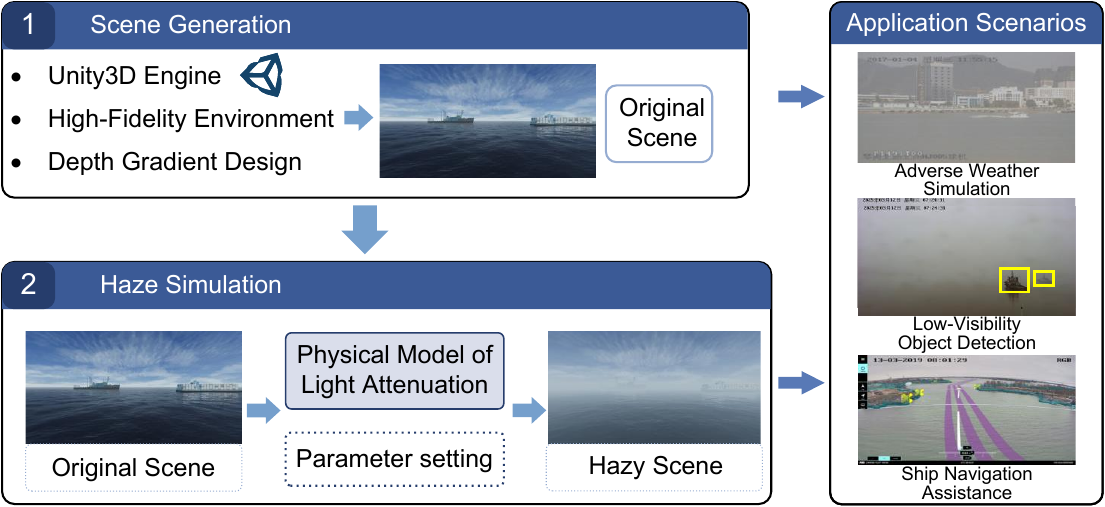}
    \caption{Flowchart of the physics-driven maritime visibility simulation and its application scenarios. It consists of high-fidelity marine environment construction, physics-based haze simulation for precise visibility distance, and the generation of multi-gradient datasets utilized for downstream tasks.}
    \label{fig:2}
\end{figure}

Building on prior work, this paper introduces MSVD, a maritime dehazing dataset with accurate annotations of visibility distance. While existing maritime dehazing datasets have significantly alleviated the shortage of paired training data for image restoration, they are primarily designed for dehazing model development and typically provide image pairs or abstract scattering coefficients rather than physically interpretable visibility measurements. To address this limitation, MSVD is specifically constructed as a visibility-oriented benchmark that enables quantitative evaluation of visibility restoration performance. In particular, MSVD provides continuous meter-level visibility annotations generated within a physics-based simulation environment, allowing image restoration performance to be directly linked to physical visibility distance. The dataset contains 12,312 simulated maritime navigation images and is generated using a controllable Unity3D-based maritime simulation platform. Unlike approaches that simply apply haze to static scenes, the proposed environment dynamically simulates multi-vessel encounters and varying camera perspectives to improve scene diversity and realism. The visibility annotations are designed to align with operational visibility thresholds commonly adopted in VTS, making the dataset particularly suitable for navigation-oriented evaluation and maritime safety research. Figure \ref{fig:2} illustrates the overall pipeline and application scenarios used for dataset construction in Unity3D.

\subsection{Scene Generation}
The physical realism of the maritime navigation simulation is achieved through the joint modeling of ocean wave dynamics and optical transport processes. To provide a robust foundation for visual enhancement and perception research, the scenario construction is optimized in the following two dimensions:

{\bf{High-Fidelity Marine Environment Dynamics: }}The water surface is modeled via a hybrid approach that reproduces realistic sea states under different environmental conditions. This formulation affords granular control over wave frequency, amplitude, and directional spreading, enabling the simulation to faithfully reproduce the intricate ocean surface morphologies driven by varying wind speeds. For optical realism, the system employs a Physically Based Rendering (PBR) workflow. We integrate Real-Time Ray Tracing (RTRT) for accurate screen-space reflections (SSR) and utilize High-Dynamic-Range (HDR) light probes to capture global illumination. This configuration ensures the physical accuracy of secondary reflections, maintaining consistency with real-world physics for the interactive light and shadow between the water surface and the vessel hull, even under typical fluctuating illumination conditions in marine environments. By accurately modeling the Bidirectional Reflectance Distribution Function (BRDF) of maritime materials, we create a marine environment with high visual fidelity and radiometric consistency, providing a reliable and deterministic foundation for subsequent visual simulation research and edge-feature analysis.

{\bf{Spatial Layout and Depth Gradient Design: }}The spatial arrangement of the navigation scene is constructed by distributing a heterogeneous set of vessels along the camera viewing direction, thereby forming a well-defined longitudinal depth structure. This layout aims to simulate the continuous visual gradient observed during maritime navigation, transitioning from the near-field to the far-field: nearby vessels maintain high-frequency structural details and sharp edges, while distant vessels gradually suffer from contrast degradation, becoming blurred and eventually merging into the atmospheric background. Based on this differentiated effect, subjective visual perception is converted into objective quantitative metrics by statistically counting the number of recognizable targets and their structural integrity under specific visibility levels. To ensure data consistency for downstream tasks such as object detection and image restoration, the visual perception data in this study is collected by a high-definition virtual camera positioned at the center of the scene.

\begin{figure*}
    \centering
    \includegraphics[width=0.95\linewidth]{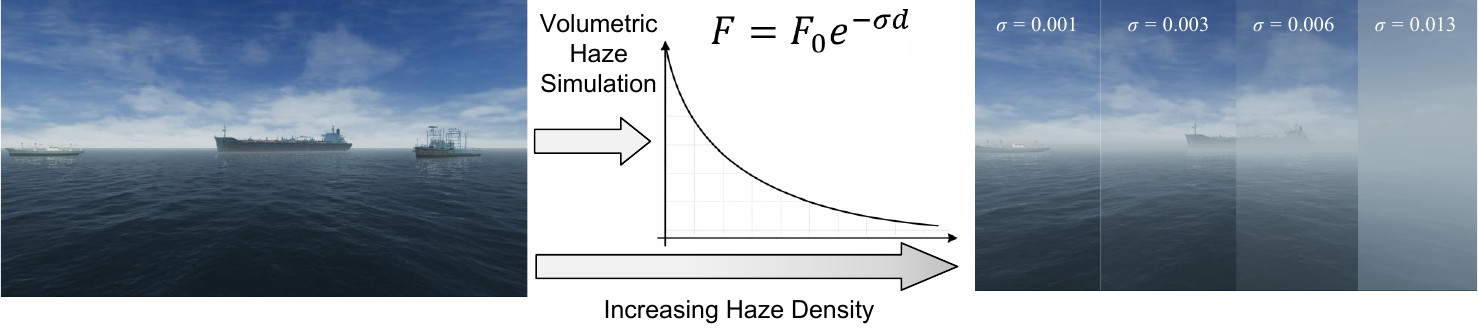}
    \caption{The haze simulation process converts a high-fidelity clear ocean scene into a multi-visibility dataset. Based on the Beer-Lambert Law, the visual contrast and structural details of the vessels gradually degrade as haze density increases.}
    \label{fig:3} 
\end{figure*}
\label{}

\subsection{Haze Simulation}
The simulation of atmospheric haze is a central component of the multi-visibility maritime navigation scenario, serving as the bridge between ideal synthetic environments and complex real-world meteorological conditions. To this end, we utilize physics-based haze simulation integrated within the Unity High Definition Render Pipeline (HDRP). By coordinating the complex interaction mechanism between its particle-based aerosol system and the dynamic lighting model, we achieved a high-fidelity haze simulation that accounts for single scattering and atmospheric absorption. This method is grounded in the physical principles of light propagation through a polydisperse medium, accurately reproducing the forward scattering and radiometric attenuation effects of haze particles on light rays. Therefore, it enables a precise dynamic simulation of maritime weather conditions ranging from subtle light haze to extreme dense haze.

{\bf{Physical Modeling of Light Attenuation: }}Atmospheric visibility is governed fundamentally by the concentration and micro-physical properties of suspended aerosols. As light propagates through a maritime hazy medium, its intensity undergoes exponential decay due to the collective scattering and absorption by micro-scale water droplets. This optical transmission loss is rigorously characterized by the classical Beer-Lambert Law:
\begin{equation}
\label{eq1}
F = F_0 e^{-\sigma d}
\end{equation}
where $ F $ and $ F_0 $ represent the observed light intensity and the incident light intensity, respectively. $\sigma$ is the extinction coefficient, and $ d $ is the distance between the target and the observation point.

{\bf{Visibility Calibration and Standardization: }}To translate this physical attenuation into a standardized visibility metric requires incorporating the visual contrast threshold. Defining $F_0$ as the ambient background intensity and $F$ as the intensity emanating from the object, the visual contrast $F/F_0$ is mathematically tied to the optical transmittance. According to the International Commission on Illumination (CIE), the minimum contrast threshold discernable by the human eye is established at 0.02. Beyond this boundary, the luminance differential between the object and its background becomes too faint for human perception. Substituting $F/F_0 = 0.02$ into Eq. (\ref{eq1}) yields the deterministic formulation linking meteorological visibility distance $V$ to the extinction coefficient $\sigma$:
\begin{equation}
\label{eq2}
V = - \frac{\log(F/F_0)}{\sigma} = \frac{3.912}{\sigma}
\end{equation}

\begin{equation}
\label{eq3}
F = F_0 e^{-\sigma D} = F_0 e^{-1}
\end{equation}
With $D=\sigma^{-1}$, the functional relationship between visibility distance $V$ and the Haze Attenuation Distance $D$ is therefore as follows:
\begin{equation}
\label{eq4}
V = 3.912*D
\end{equation}

By modulating the Haze Attenuation Distance parameter $D$, synthetic maritime navigation images across a spectrum of visibility conditions are generated. The computational pipeline for haze simulation is delineated in Figure \ref{fig:3}.

To guarantee operational relevance and offer practical guidance for maritime safety, the dataset aligns its visibility levels with the International Regulations for Preventing Collisions at Sea (COLREGs). Based on these standards, visibility distances are systematically classified into six distinct grades: Very Dense Haze (under 50 m), Dense Haze (under 200 m), Moderate Haze (under 500 m), Light Haze (under 1000 m), Thin Haze (under 2000 m), and Poor Visibility (under 4000 m).

\begin{figure*}[t]
    \centering
   \includegraphics[width=0.95\linewidth]{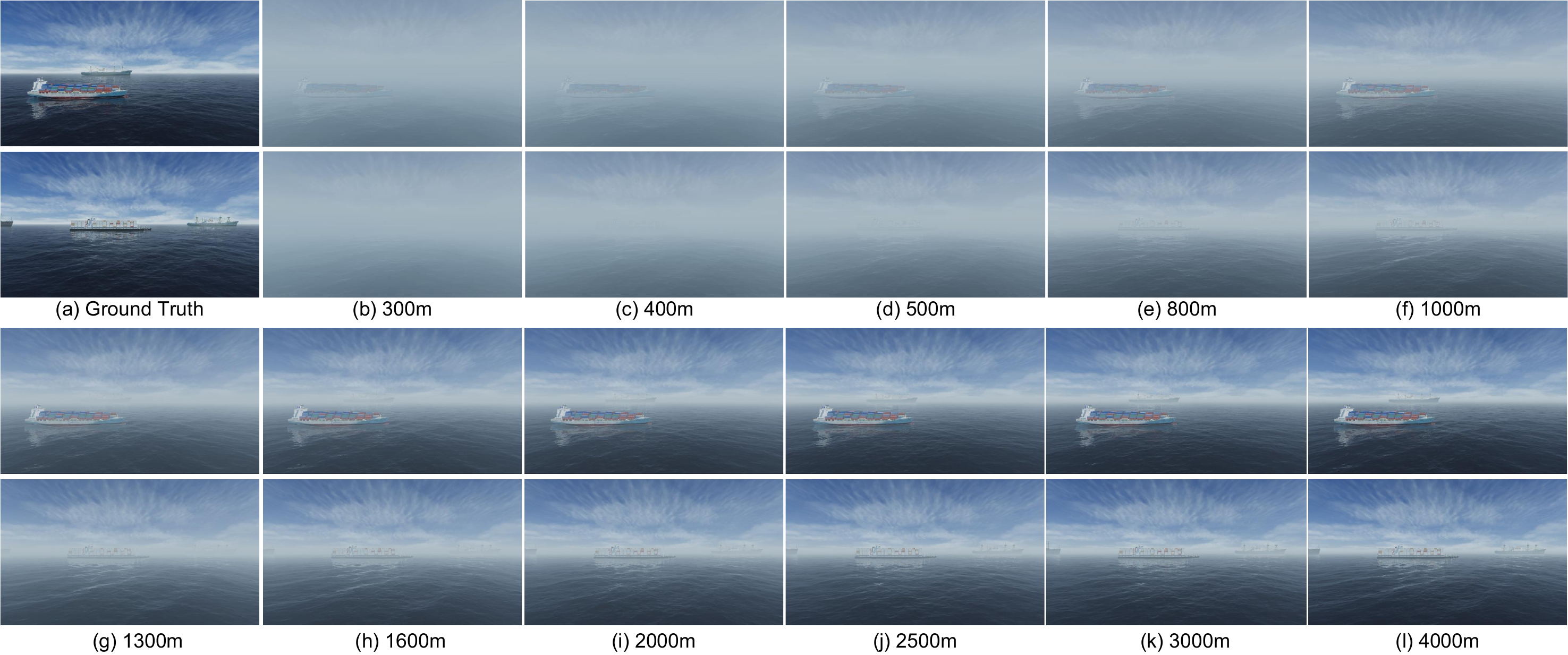}
    \caption{Representative samples from the simulated maritime dataset, including ground truth images and annotated visibility distance levels ranging from 300 m to 4000 m. The visibility distance values denote the effective observation distance under haze conditions, where lower values represent severely restricted visibility distance and higher values approach near-clear maritime environments. The simulated images demonstrate the gradual attenuation of target contrast and edge structures as visibility distance decreases, offering a physically interpretable basis for visibility evaluation.}
    \label{fig:4}
\end{figure*}

To establish a continuous visibility gradient based on the aforementioned categorization, we selected 300 m, 500 m, 1000 m, 2000 m, and 4000 m as the primary baseline levels. To further densify these intervals, additional visibility levels were interpolated between adjacent baselines (e.g., inserting 1300 m and 1600 m between the 1000 m and 2000 m anchors). This systematic construction yields a multi-level, high-precision simulation dataset that comprehensively spans a wide range of visibility conditions, serving as a robust foundation for subsequent performance evaluations. Representative images illustrating these diverse visibility levels are presented in Figure \ref{fig:4}.

\section{Visibility Quantification Method Based on Object Detection}
Maritime haze significantly degrades visibility distance, corrupting the features required for object detection, which consequently leads to a reduction in object detection accuracy \citep{ZHANG2025121447}. Therefore, image visibility is inherently linked to object detection precision. Task driven evaluation has been widely adopted in computer vision to assess image enhancement performance through downstream perception tasks. Building upon this concept, the proposed framework extends task performance evaluation to visibility quantification in maritime environments. Unlike conventional task driven metrics that report only detection accuracy, the proposed metric establishes a calibrated relationship between object detection performance and physical visibility distance using the MSVD benchmark. Through this calibration process, improvements in object detection accuracy can be translated into physically interpretable visibility gains. Furthermore, by aligning the estimated visibility distance with maritime operational visibility thresholds (Table \ref{tab:visibility}), the proposed framework enables image enhancement performance to be evaluated from the perspective of navigational safety rather than solely through conventional image quality metrics.

\subsection{Object Detection Models}
The reliability of maritime autonomous navigation depends on the accurate detection and classification of heterogeneous vessels under diverse meteorological conditions. In this study, we adopt representative models from two fundamentally different object detection paradigms: single-stage and two-stage detectors. These paradigms exhibit distinct trade-offs in terms of computational efficiency, localization accuracy, and robustness, which are highly relevant to maritime engineering applications.

Single-stage detectors formulate object detection as a unified regression task, directly predicting object locations and categories in a single forward pass\citep{chen2023orientation}. Their streamlined structure enables high inference speed and reduced computational burden, making them particularly suitable for real-time maritime surveillance and onboard intelligent systems where hardware resources and response latency are critical constraints. However, due to the absence of an explicit region refinement stage, their localization accuracy and robustness may degrade more noticeably under severe haze conditions where feature quality is compromised. In contrast, two-stage detectors adopt a cascaded framework in which candidate regions are first generated and then refined through dedicated classification and localization modules. This design generally achieves higher localization precision and stronger robustness in complex or low-contrast scenes, albeit at the cost of increased computational complexity. Such architectures are often more suitable for shore-based monitoring systems or offline analysis scenarios where higher accuracy is prioritized over real-time performance.

\subsection{Visibility Distance Estimation for Image Dehazing}
\begin{figure*}
    \centering
    \includegraphics[width=0.95\linewidth]{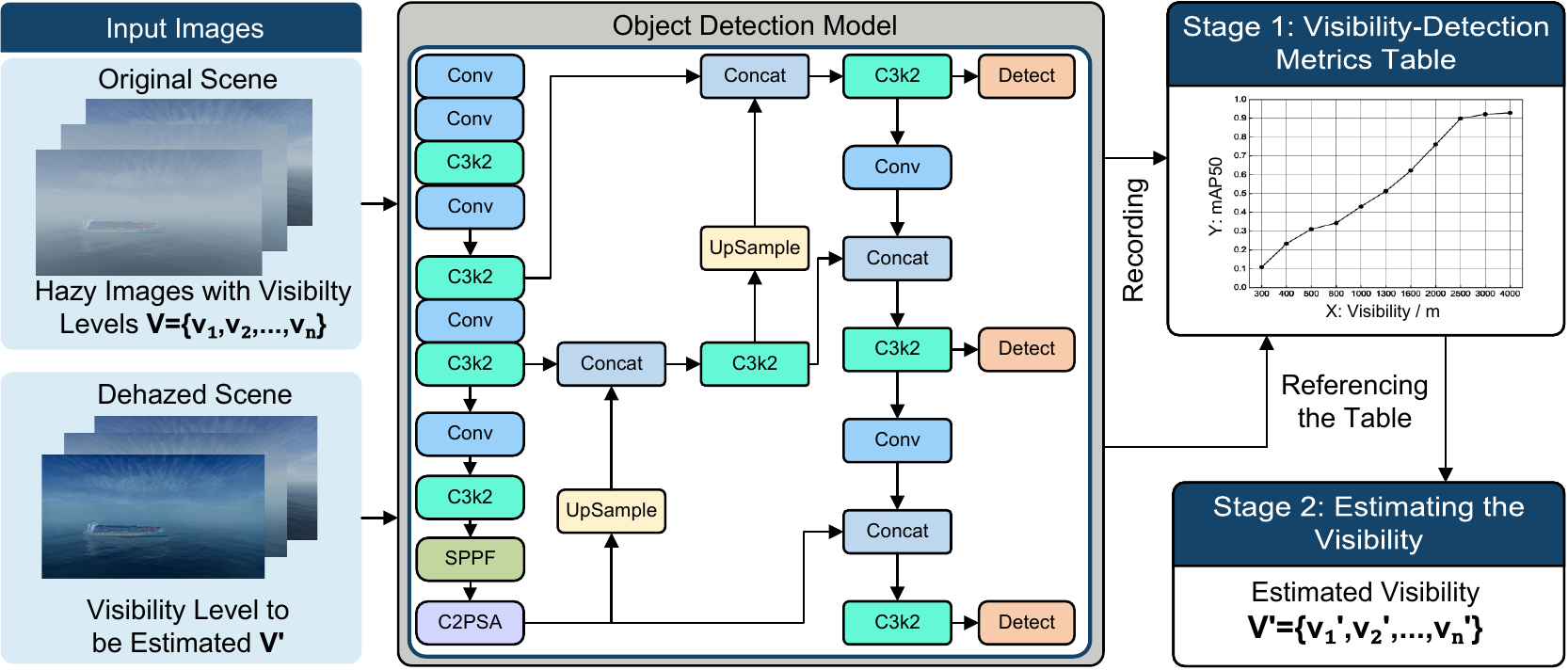}
    \caption{Flowchart of the proposed visibility quantification method based on object detection accuracy correlation. The process comprises two primary stages: 1) Construction of a Visibility-Detection Accuracy Correlation Table ($I$) using original hazy images with preset visibility distance levels; 2) Estimation of the dehazed visibility distance ($V'$) by inputting dehazed images into the same detection model to obtain accuracy metrics ($I'$), followed by a linear interpolation process referencing the established correlation table. This framework transforms discrete detection performance into a continuous and objective visibility metric.}
    \label{fig:5} 
\end{figure*}

The proposed quantification method leverages the performance metrics of these models as a bridge for atmospheric visibility. The specific process, which bridges the gap between low-level image restoration and high-level semantic understanding, is illustrated in Figure \ref{fig:5}. To transform abstract detection performance into a concrete measure of environmental visibility distance, we implement a two-stage evaluation framework.

{\bf{Stage 1 Establish the Visibility-Detection Baseline: }} The proposed method begins by predefining a set of visibility levels, $V = {v_1, v_2, \dots, v_n}$, based on the MSVD. To establish a calibrated relationship between physical visibility and semantic perception performance, a vessel detection model pre-trained on large-scale real-world maritime datasets is employed. For each visibility level, the detector evaluates the complete image set associated with that visibility condition, and the corresponding mean Average Precision (mAP50) is calculated. Since mAP50 is a dataset-level metric that jointly reflects localization and classification performance, it provides a statistically robust representation of the detector’s perception capability under a given visibility condition. Based on the resulting set of mAP50 values, $\mathbf{I} = {I_1, I_2, \dots, I_n}$, a visibility–detection metric table is constructed. This table serves as a calibrated benchmark that links objective meteorological visibility levels to the detector's semantic perception performance.

{\bf{Stage 2 Quantification via Continuous Interpolation: }} To evaluate a dehazing method, all hazy images corresponding to a specific visibility level are first processed by the selected restoration model. The same YOLO11 detector is then applied to the entire dehazed image set, yielding a restored detection accuracy $I'_n$ in terms of mAP50. It should be noted that $I'_n$ does not correspond to an individual image but rather represents the statistical detection performance of the complete image set after dehazing. To overcome the limitation of discrete visibility categories, a linear interpolation strategy is employed. For any restored accuracy value $I'_n$ falling within the interval $[I_j, I_{j+1}]$, the corresponding visibility distance $v'_n$ is calculated as:
\begin{equation}
\label{eq5}
v'_n = v_j + \frac{I'_n - I_j}{I_{j+1} - I_j} * (v_{j+1} - v_j)
\end{equation}

The numerical value derived from Eq. (\ref{eq5}) provides a continuous metric for visibility improvement. Piecewise linear interpolation is chosen over higher-order curve fitting to preserve the monotonic relationship. Since object detection accuracy correlates monotonically with physical visibility, higher-order methods risk introducing non-physical local oscillations. Given the dense baseline intervals in MSVD, this linear approach avoids overfitting and strictly aligns with maritime safety regulations, which prioritize discrete visibility thresholds over absolute precision.

Furthermore, the generalizability of the proposed metric across different detection architectures was examined through extensive cross-validation experiments (detailed in Section 5.3). The highly consistent visibility restoration trends observed across different detectors indicate that the framework can provide stable evaluation results while effectively linking image restoration performance to navigational safety considerations.

The practical validity of the proposed calibration mapping was established through a carefully controlled experimental protocol conducted on the MSVD. First, scene consistency was maintained to minimize the influence of confounding variables. Identical 3D navigational scenes comprising the exact same vessel spatial coordinates, dynamic trajectories, and camera perspectives—were rendered in the Unity3D engine, with only the physical atmospheric scattering coefficient systematically adjusted to designated baseline levels. Second, the mAP50 baselines were determined using the full corpus of images at each distinct visibility level, thereby ensuring maximum statistical reliability. Finally, all bounding box ground truths were annotated after image collection. This precise annotation process guarantees high-quality labels that strictly align with standard evaluation protocols, providing an objective foundation for accurate mAP50 computation.

\label{}
\section{Experiment Results and Analysis}
To comprehensively validate whether existing dehazing evaluation criteria can truly reflect visibility distance improvement in maritime navigation scenarios, and to demonstrate the practical significance of the proposed visibility distance quantification metric, a series of systematic experiments is conducted in this paper. Specifically, representative dehazing methods are first evaluated under multi-visibility conditions using commonly adopted full-reference and no-reference image quality metrics, revealing their inherent limitations when applied to navigation-oriented visibility distance assessment. Subsequently, a task-driven visibility distance quantification framework, integrated with object detection performance, is established and validated, enabling a direct and physically meaningful interpretation of dehazing efficacy. Finally, robustness evaluations across multiple detection architectures are performed to further ascertain the reliability and generalization capability of the proposed metric. It is worth noting that during this evaluation phase, the object detectors operate in a deterministic inference mode on the MSVD. Because this pure testing process involves no stochastic training optimization, running the pipeline multiple times does not produce meaningful numerical variance. Instead, the stability and reliability of the proposed metric are demonstrated by its highly consistent assessment trends across all 11 distinct visibility levels and 4 detector backbones.

\label{}

\subsection{Benchmarks and Traditional Metric}
\begin{figure*}
    \centering
    \includegraphics[width=0.95\linewidth]{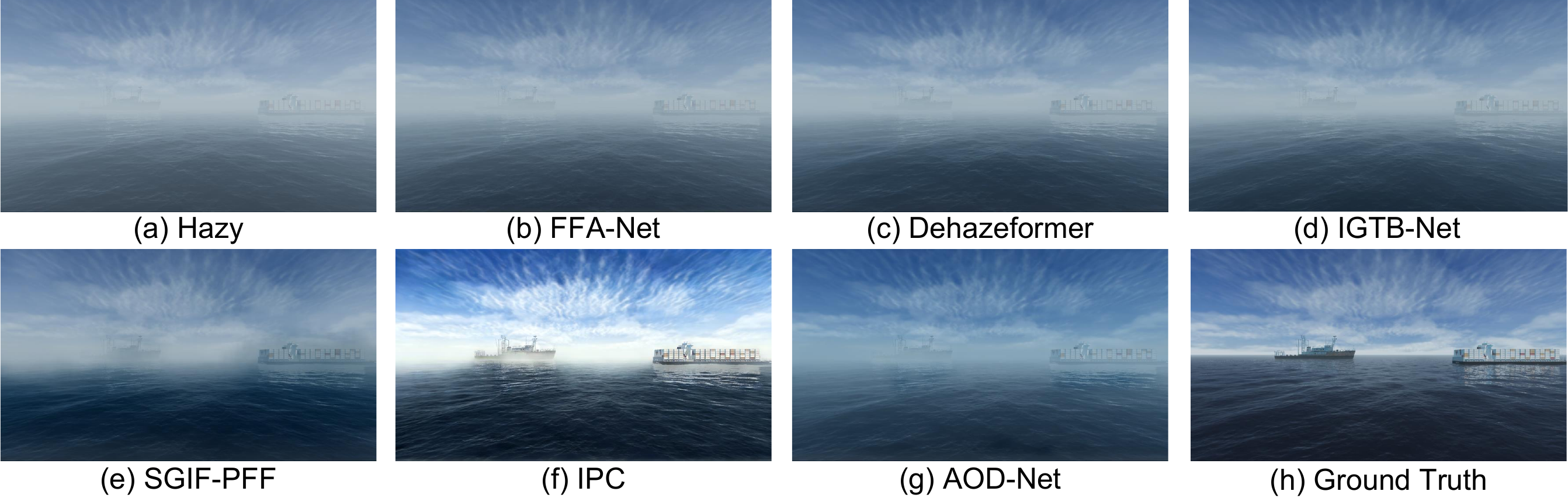}
    \caption{Visual comparison of different dehazing results. To provide comparative benchmarks, six representative methods—including FFA-Net, DehazeFormer, IGTB-Net, SGIF-PFF, IPC and AOD-Net were evaluated using official pre-trained implementations. The results demonstrate the diverse performance of various technical approaches in restoring visibility and structural details from the original hazy maritime images.}
    \label{fig:6}
\end{figure*}

To evaluate the performance of the dehazing methods and provide comparative benchmarks for the quantitative visibility distance assessment metric proposed in this paper, six representative dehazing methods were selected for experimentation. The selected methods comprise several distinct technical approaches, including: FFA-Net \citep{qin2020ffa}, DehazeFormer \citep{song2023vision}, IGTB-Net \citep{liu2024interaction}, SGIF-PFF \citep{bai2022self}, IPC \citep{fu2025iterative}, and AOD-Net \citep{li2017aod}. Figure \ref{fig:6} illustrates the visual comparisons of the dehazing results produced by these candidate methods.

{\bf{FFA-Net: }}FFA-Net is an end-to-end framework that restores haze-free images without explicitly estimating the parameters of the atmospheric scattering model. Its main contribution is a Feature Attention (FA) module that combines channel attention and pixel attention, which enables the network to assign different weights to feature channels and spatial locations according to the distribution of haze.

{\bf{Dehazeformer: }}DehazeFormer is a Transformer-based architecture designed for single-image dehazing. In contrast to standard Vision Transformers designed for high-level tasks, DehazeFormer introduces several architectural optimizations. It adopts a modified normalization layer to preserve patch relationships and utilizes ReLU activation to facilitate feature recovery. It applies a shifted window partitioning strategy with reflection padding to ensure consistent window sizes.

{\bf{IGTB-Net: }}IGTB-Net proposes a novel dual-branch architecture that leverages the complementary characteristics of CNNs and Transformers for image dehazing. To mitigate feature redundancy caused by overlapping extraction across the two branches, the model introduces a downsampling operation before the Transformer branch to efficiently capture the global context. This extracted global information is then utilized to interactively guide the CNN branch via a Channel and Pixel Attention (CPA) block, allowing the network to focus precisely on effective local details while maintaining overall global consistency.

{\bf{SGIF-PFF: }}SGIF-PFF explores guidance information from the input hazy image itself. The framework includes a Deep Pre-dehazer that produces an intermediate reference image with clearer structural information. To leverage this guidance, it employs a Progressive Feature Fusion module that integrates features from both the original hazy image and the reference image in a step-by-step manner.

{\bf{IPC: }}IPC is a framework for real-world image dehazing that incorporates codebook priors obtained from a pre-trained VQGAN. Instead of relying on one-shot decoding, IPC introduces an iterative predictor-critic mechanism. The framework contains a Code-Predictor that estimates discrete codes from hazy image features and a Code-Critic that evaluates the reliability of these predictions.

{\bf{AOD-Net: }}AOD-Net is an early end-to-end dehazing model based on convolutional neural networks that reformulates the classical atmospheric scattering model. Rather than estimating the transmission map and atmospheric light separately, which can lead to error accumulation, AOD-Net combines these parameters into a single variable ($K$). This formulation enables the network to generate haze-free images directly through a lightweight architecture.

\label{}

\begin{figure*}[t!]
\centering
\includegraphics[width=0.98\linewidth]{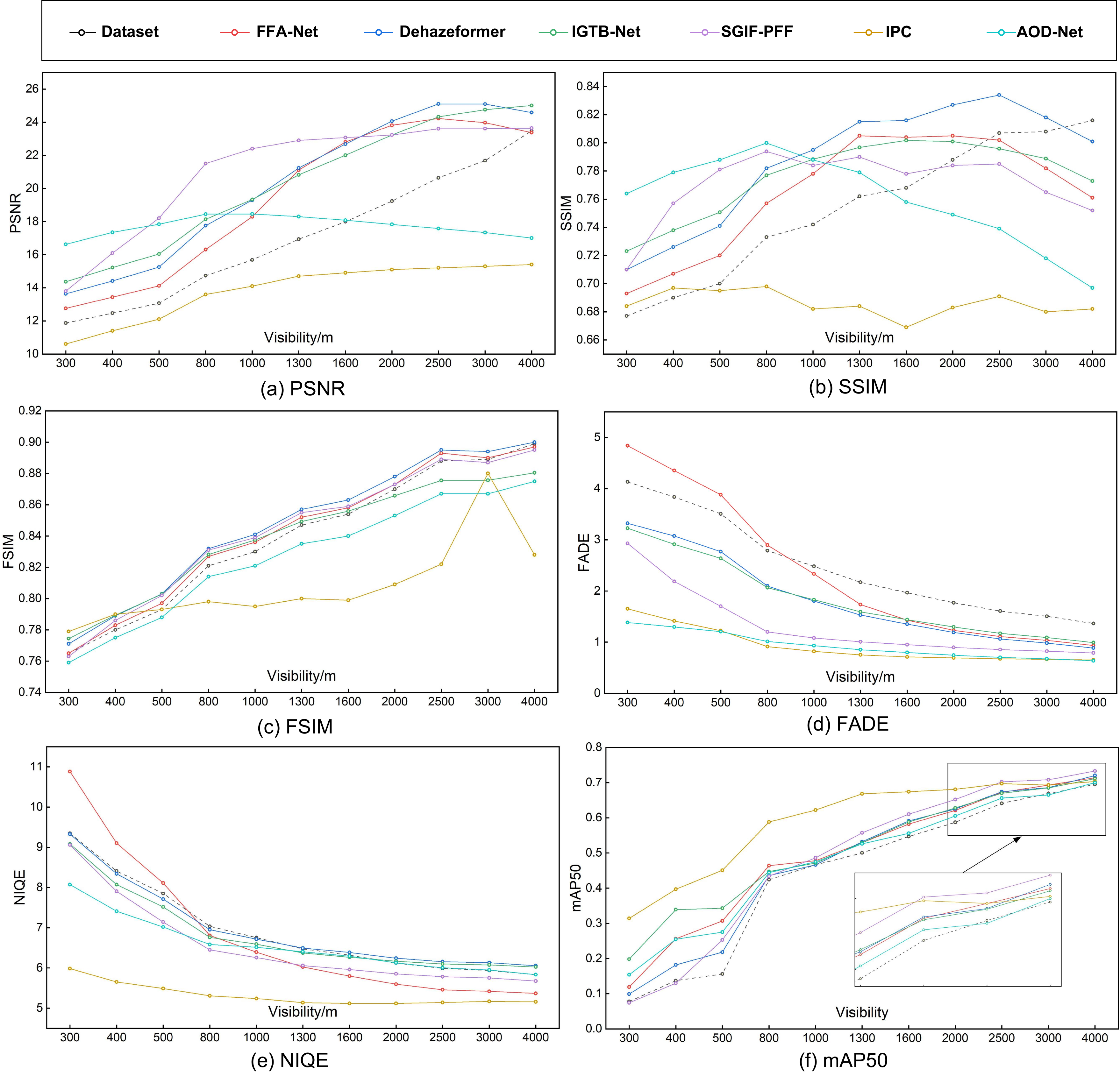}
\caption{Performance of different dehazing models under varying visibility distance conditions. From top-left to bottom-right are the results for (a) PSNR, (b) SSIM, (c) FSIM, (d) FADE, (e) NIQE, and (f) mAP50 across various dehazing methods. The results indicate that FR metrics (a)-(c) fail to match the visibility improvement trend. While the NR metrics (FADE and NIQE) align with the visibility trend, they show significant variance across different methods. Conversely, mAP50, serving as an object detection-based metric, demonstrates a trend consistent with visibility gains and a stable distribution across methods, effectively reflecting the improvement in visibility after dehazing.}
\label{fig:7}
\end{figure*}

In this paper, the original images under varying visibility distance conditions are first processed using dehazing methods, after which commonly adopted dehazing evaluation metrics are computed. The selected metrics include FR metrics: Peak Signal-to-Noise Ratio (PSNR), Structural Similarity Index Measure (SSIM), and Feature Similarity Index Measure (FSIM); NR metrics: Fog-density, AC-distribution-based Dehazing Evaluator (FADE), and Natural Image Quality Evaluator (NIQE). The traditional metric data for the dehazing results under various visibility distance levels is shown in Figure \ref{fig:7}.

The results obtained from these established metrics provide an initial assessment of the method's efficacy, which will further be compared with the novel quantitative visibility assessment metric proposed in this study. This comparative analysis is intended to highlight the unique value of our proposed metric, specifically within the challenging maritime navigation scenario.

\begin{figure*}[t!]
\centering
\includegraphics[width=0.9\linewidth]{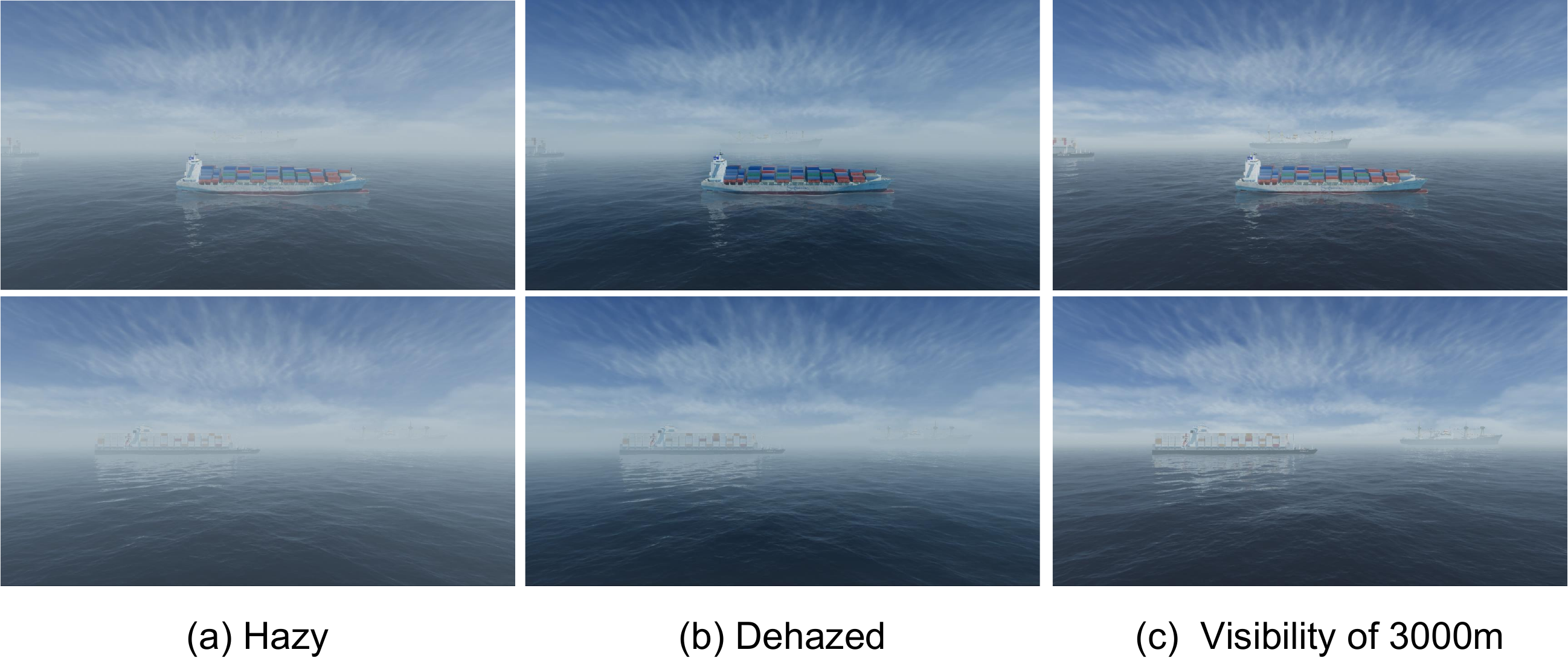}
\caption{Inconsistency between no-reference visibility estimation and dataset visibility levels. (a) Hazy image with visibility of 1300 m in the dataset. (b) Dehazed result of (a), for which the no-reference metric predicts a visibility value close to 3000 m. (c) Image from the dataset with visibility of 3000 m. Although the no-reference metric indicates a substantial visibility improvement after dehazing, visual comparison with the dataset 3000 m scene shows that the actual restoration of scene transparency and distant vessel details remains noticeably inferior. This example indicates that variations in no-reference metric values do not necessarily correspond to equivalent annotated visibility conditions.}

\label{fig:8}
\end{figure*}

{\bf{FR metrics: }}Figure \ref{fig:7} (a), (b), and (c) present the FR metrics performance of different dehazing models. Overall, in most test conditions, the improvement in image visibility exhibits a positive correlation with the optimization of objective evaluation metrics, and the majority of these metrics are enhanced after dehazing, which is consistent with the initial purpose of employing a dehazing model. However, the optimal performance value for the metrics typically appears in moderate haze scenarios rather than in scenes with the theoretically highest visibility. Specifically, while the dehazing process significantly clarifies vessel contours and surface textures, it simultaneously introduces subtle color shifts and saturation biases. These discrepancies adversely impact pixel-based metrics such as PSNR, which prioritize absolute color fidelity. Furthermore, in a few test scenarios, the metrics of the dehazed images are even lower than those of the original hazy images, which is related to the dehazing model changing the image's tone or style. However, the comparison of dehazing results shown in Figure \ref{fig:6} under haze conditions confirms that the dehazing models effectively suppress the haze covering the target surface and distant areas.

The result reveals a disconnect between traditional FR metrics and their ability to assess the actual enhancement of visibility for maritime navigation, because they tend to penalize distortions introduced during the dehazing process but fail to accurately reflect the substantial value contributed by the processing results to the actual detection range and navigational safety at sea.

{\bf{NR metrics: }}We also conducted a quantitative analysis of the NR metrics for dehazed images under different visibility conditions, with the results shown in Figure \ref{fig:7} (c) and (d). The FADE results indicate that the metric is optimized after dehazing, we found that the metric value range of the original images cannot adequately cover the range of the dehazed images. This causes the FADE value for the original low-visibility hazy images, after processing by the dehazing method, to leap directly to a level close to the metrics of high-visibility images. Figure \ref{fig:8} displays the dehazing result for 1300 m and the original image with 3000 m visibility. Visually, the dehazed image only removes most of the haze but fails to truly and clearly display the distant target vessel, thus failing to achieve a significant improvement in actual visibility. Therefore, the improvement observed in the FADE metric severely deviates from the actual contribution of the dehazing method to visibility enhancement. Conversely, the NIQE metric exhibits relative insensitivity to changes in visibility, suggesting that it captures image naturalness rather than haze concentration. The phenomena of the FADE metric jump and NIQE lack of responsiveness collectively demonstrate that the numerical change of these NR metrics cannot effectively reflect the actual improvement in visibility, exposing the limitations of NR metrics in practical nautical applications.

\newcolumntype{Y}{>{\centering\arraybackslash}X}
\begin{table*}[t!]
\centering
\caption{Quantitative comparison of six dehazing methods under different visibility levels using the YOLO11 detection framework. 
The upper block reports detection performance (mAP50), while the lower block presents the corresponding visibility values (New Vis) quantified by the proposed visibility estimation method. 
'Baseline' refers to the reference mapping derived from the original hazy images; '-' indicates that the quantified visibility over 4000 m.}
\label{tab2}
\small
\setlength{\tabcolsep}{4pt}
\renewcommand{\arraystretch}{1.15}
\begin{tabularx}{\textwidth}{l *{11}{Y}}
\toprule
\textbf{Model / Vis (m)} & 300 & 400 & 500 & 800 & 1000 & 1300 & 1600 & 2000 & 2500 & 3000 & 4000 \\
\midrule
\multicolumn{12}{c}{\textbf{mAP50}} \\
\midrule
\textbf{Baseline} & \textbf{0.078} & \textbf{0.137} & \textbf{0.156} & \textbf{0.424} & \textbf{0.466} & \textbf{0.500} & \textbf{0.547} & \textbf{0.587} & \textbf{0.641} & \textbf{0.669} & \textbf{0.695} \\
AOD-Net \citep{li2017aod} & 0.154 & 0.255 & 0.275 & 0.444 & 0.475 & 0.526 & 0.556 & 0.605 & 0.656 & 0.665 & 0.700 \\
FFA-Net \citep{qin2020ffa} & 0.119 & 0.256 & 0.307 & 0.464 & 0.478 & 0.529 & 0.582 & 0.621 & 0.672 & 0.693 & 0.714 \\
SGIF-PFF \citep{bai2022self} & 0.074 & 0.130 & 0.253 & 0.434 & 0.486 & 0.557 & 0.610 & 0.652 & 0.702 & 0.708 & 0.733 \\
DehazeFormer \citep{song2023vision} & 0.099 & 0.182 & 0.218 & 0.436 & 0.466 & 0.532 & 0.591 & 0.625 & 0.674 & 0.686 & 0.720 \\
IGTB \citep{liu2024interaction} & 0.198 & 0.339 & 0.343 & 0.447 & 0.471 & 0.529 & 0.588 & 0.628 & 0.670 & 0.685 & 0.711 \\
IPC \citep{fu2025iterative} & 0.314 & 0.397 & 0.451 & 0.588 & 0.622 & 0.668 & 0.674 & 0.681 & 0.697 & 0.693 & 0.703 \\
\midrule
\multicolumn{12}{c}{\textbf{New Vis (m)}} \\
\midrule
AOD-Net \citep{li2017aod}
& \viscell{489}{+189}
& \viscell{611}{+211}
& \viscell{633}{+133}
& \viscell{895}{+95}
& \viscell{1079}{+79}
& \viscell{1466}{+166}
& \viscell{1690}{+90}
& \viscell{2167}{+167}
& \viscell{2768}{+268}
& \viscell{2929}{-71}
&  - \\

FFA-Net \citep{qin2020ffa}
& \viscell{369}{+69}
& \viscell{612}{+212}
& \viscell{669}{+169}
& \viscell{990}{+190}
& \viscell{1106}{+106}
& \viscell{1485}{+185}
& \viscell{1950}{+350}
& \viscell{2315}{+315}
& \viscell{3115}{+615}
& \viscell{3923}{+923}
& -\\

SGIF-PFF \citep{bai2022self}
& \viscell{300}{+0}
& \viscell{388}{-12}
& \viscell{609}{+109}
& \viscell{848}{+48}
& \viscell{1176}{+176}
& \viscell{1700}{+400}
& \viscell{2213}{+613}
& \viscell{2696}{+696}
& -
& -
& - \\

DehazeFormer \citep{song2023vision}
& \viscell{336}{+36}
& \viscell{529}{+129}
& \viscell{569}{+69}
& \viscell{857}{+57}
& \viscell{1000}{+0}
& \viscell{1504}{+204}
& \viscell{2037}{+437}
& \viscell{2352}{+352}
& \viscell{3192}{+692}
& \viscell{3654}{+654}
& - \\

IGTB \citep{liu2024interaction}
& \viscell{547}{+247}
& \viscell{705}{+305}
& \viscell{709}{+209}
& \viscell{910}{+110}
& \viscell{1044}{+44}
& \viscell{1485}{+185}
& \viscell{2009}{+409}
& \viscell{2380}{+380}
& \viscell{3038}{+538}
& \viscell{3615}{+615}
& - \\

IPC \citep{fu2025iterative}
& \viscell{677}{+377}
& \viscell{770}{+370}
& \viscell{929}{+429}
& \viscell{2009}{+1209}
& \viscell{2324}{+1324}
& \viscell{2982}{+1682}
& \viscell{3192}{+1592}
& \viscell{3462}{+1462}
& -
& \viscell{3923}{+923}
& -\\
\bottomrule
\end{tabularx}
\end{table*}

We also establish the relationship between mAP50 and the corresponding visibility, as illustrated in Figure \ref{fig:7} (f). Compared to the aforementioned FR and NR metrics, the mAP50-Visibility curve demonstrates a much higher degree of consistency with the objective changes in meteorological visibility. Although a few data points exhibit minor fluctuations or outliers under extreme conditions, the overall trend of this metric aligns more closely with the actual perceptual capacity in maritime scenarios. Notably, the proposed task-driven metric effectively captures its substantial contribution to visibility restoration. This suggests that while traditional metrics might over-penalize the specific style shifts or artifacts, our approach is able to capture its effectiveness in improving object detectability and navigational safety.

In summary, when applied to the evaluation of dehazing methods for low-visibility maritime images, neither traditional FR metrics, nor NR metrics, are capable of accurately reflecting the actual needs of maritime navigation. They both exhibit a disconnect from practical requirements and lack the ability to provide a quantitative assessment of dehazing visibility, which is crucial for maritime safety.

\subsection{Evaluation of Visibility Quantification}

Beyond traditional image quality assessment, enabling quantitative visibility estimation requires a metric that directly relates image restoration performance to visibility distance. In response to this need, this paper proposes an object detection-based metric for quantifying visibility in dehazing scenarios. By leveraging the object detection performance indicator mAP, the proposed metric correlates image restoration quality with measurable visibility improvement, thereby translating the abstract notion of image quality enhancement into a quantity with explicit physical significance. 

In the experimental framework, YOLO11 is employed as the benchmark detection model, having been pre-trained on the Singapore Maritime Dataset. Utilizing the images and precise visibility labels from our dataset, we analyze the relationship between visibility levels and the object detection performance metric, establishing a reference mapping as summarized in Table \ref{tab2}.

\begin{figure*}[t!]
\centering
\includegraphics[width=0.97\linewidth]{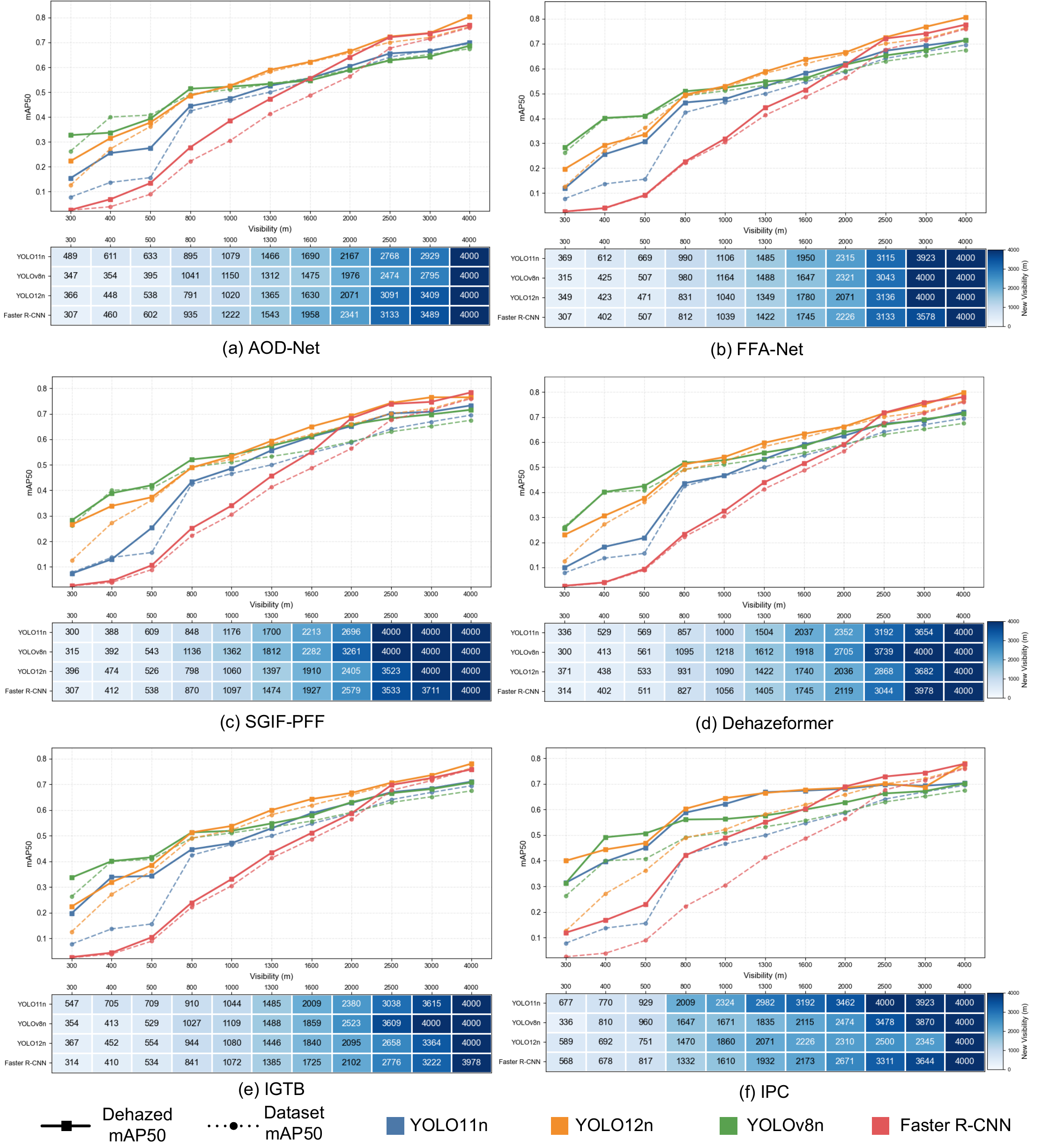}
\caption{Performance evaluation of various dehazing methods across different object detection models. Figure (a)–(f) corresponds to a specific dehazing method: (a) FFA-Net, (b) Dehazeformer, (c) IGTB-Net, (d) SGIF-PFF, (e) IPC, and (f) AOD-Net. Within each figure, the upper line graph illustrates the detection performance, where solid lines represent the dehazed mAP50 and dashed lines denote the baseline mAP50 from the original hazy images; different colors indicate distinct object detection models. The lower heatmap in each figure displays the corresponding calculated post-dehazing visibility estimated for each detection algorithm.}
\label{fig:9}
\end{figure*}

Compared with traditional image quality metrics such as PSNR and SSIM, the proposed metric directly evaluates the substantial contribution of image processing results to visibility rather than pixel-level distortion. This visibility-oriented design enables evaluation results to directly serve navigational safety decisions in marine traffic systems, effectively overcoming the limitations of traditional metrics. With object detection performance as its core, the proposed metric achieves a task‑driven evaluation of dehazing methods. Therefore, the proposed metric ensures that while the dehazing model enhances visual effects, it also improves the accuracy and robustness of downstream detection tasks.

Using the proposed metric, we evaluated the performance of six selected dehazing methods under different visibility levels and quantitatively analyzed their dehazing visibility. The experimental results are shown in Table \ref{tab2}.

A representative example is provided below to illustrate both the quantitative mechanism and the practical utility of the proposed metric. Consider an original hazy scene with a physical visibility of 800 m, where the baseline detection mAP50 is 0.424. According to standard Vessel Traffic Services regulatory thresholds, this environment is classified under the "Restricted Navigation" category, which requires authorities to implement mandatory speed limits and traffic control. When processing this scene, different dehazing methods yield varying detection performances, which our metric maps directly to actionable safety decisions using Eq. (\ref{eq5}). For instance, one top-performing method improves the mAP50 to 0.588. Interpolating against the established baseline points, the estimated New Visibility ($V_{new}$) surges to 2009 m. By crossing the 2000 m safety boundary, the operational status is officially upgraded to "Minor Restrictions," providing a clear quantitative justification for VTS operators to lift traffic constraints and optimize waterway efficiency. In contrast, another evaluated method yields a minor mAP50 improvement to 0.444, which translates to an estimated $V_{new}$ of only 895 m. Because this fails to break through the 1000 m threshold, the system is forced to maintain the "Restricted Navigation" status. This case study demonstrates how the proposed framework successfully translates abstract image enhancement gains into regulation-aligned navigational decisions.

\subsection{Robustness and Deployment Feasibility Analysis}
We extended our cross-validation to three additional representative detectors widely utilized in maritime surveillance: YOLOv8n, YOLO12n, and Faster R-CNN. As illustrated in Figure \ref{fig:9}, which presents the mAP50 and the corresponding New Visibility for six dehazing methods, the experimental results demonstrate a generally consistent trend across different detection architectures. While the absolute baseline and restored performance (dashed and solid lines) vary with the specific detector, the derived visibility metrics (heatmap) follow a similar distribution pattern. Although minor numerical fluctuations exist among the different backbones, the relative performance ranking of the dehazing methods remains largely stable across Figure \ref{fig:9} (a)–(f). This indicates that the proposed visibility metric is capable of reflecting the overall trend of visibility restoration, suggesting its potential as a robust evaluation benchmark that is not overly dependent on a specific detection model.

Furthermore, to properly interpret the observed variance in absolute estimated visibility values among these detectors, it is essential to note that this detector-dependency is algorithmically reasonable. Different network architectures possess distinct feature-extraction capacities, which inherently alter the initial baseline mapping curve between the mAP and physical visibility. Consequently, the proposed framework should be regarded as a calibration-based visibility evaluation method, where each detector establishes its own visibility–accuracy mapping through an initial calibration process. In practical deployment, a selected object detector must first generate its specific baseline mapping curve. Once this initial calibration is established, the framework reliably translates detection accuracy into meaningful visibility distances for that specific system, while continuing to consistently evaluate the performance gains of different dehazing methods.

Beyond robustness, the practical deployment feasibility of the proposed framework is also an important consideration for maritime applications. For practical shipborne deployment, computational efficiency and real-time performance are critical factors. The computational framework of the proposed metric is inherently optimized for resource-constrained edge devices. First, the most computationally intensive phase, which involves establishing the visibility–detection metric table, is executed entirely offline. In a real-world deployment scenario, the onboard system would merely perform lightweight table lookups and linear interpolation, consuming negligible resources. Second, since object detection is already a fundamental component of modern intelligent maritime surveillance, our metric can effectively reuse these existing inference pipelines without requiring an additional dedicated heavy network. Considering that meteorological visibility in maritime environments changes gradually, periodic sampling would be sufficient for navigational safety, drastically reducing the overall computational burden over time. 

To further evaluate deployment feasibility, Table \ref{tab:computational_cost} summarizes the computational complexity and inference speed of the representative detection models used in this study. The single-stage detectors demonstrate clear advantages for edge deployment. Compared to the two-stage Faster R-CNN, these lightweight models require significantly fewer parameters, lower computational overhead (FLOPs), and a much smaller memory footprint, while achieving substantially higher inference speeds (FPS). This structural efficiency confirms the theoretical capability of single-stage architectures to satisfy real-time requirements on standard maritime edge computing devices.

\begin{table}[t!]
\centering
\renewcommand{\arraystretch}{1.2}

\caption{Quantitative comparison of computational overhead and inference speed among evaluated object detection architectures.}
\label{tab:computational_cost}
\resizebox{\columnwidth}{!}{
\begin{tabular}{lccccc}
\toprule
\textbf{Detection Model} & \textbf{Parameters (M)} & \textbf{FLOPs (G)} & \textbf{Model Size (MB)} & \textbf{Inference Time (ms)} & \textbf{FPS} \\
\midrule
YOLOv8n & 3.2 & 8.9 & 6.3 & 5.7 & 174 \\
YOLO11n & 2.6 & 6.6 & 5.4 & 7.7 & 129 \\
YOLO12n & 2.6 & 6.7 & 5.3 & 12.1 & 83 \\
Faster R-CNN & 41.4 & 250.8 & 158.2 & 32.5 & 31 \\
\bottomrule
\end{tabular}
}
\end{table}

\begin{figure*}[t!]
\centering
\includegraphics[width=0.97\linewidth]{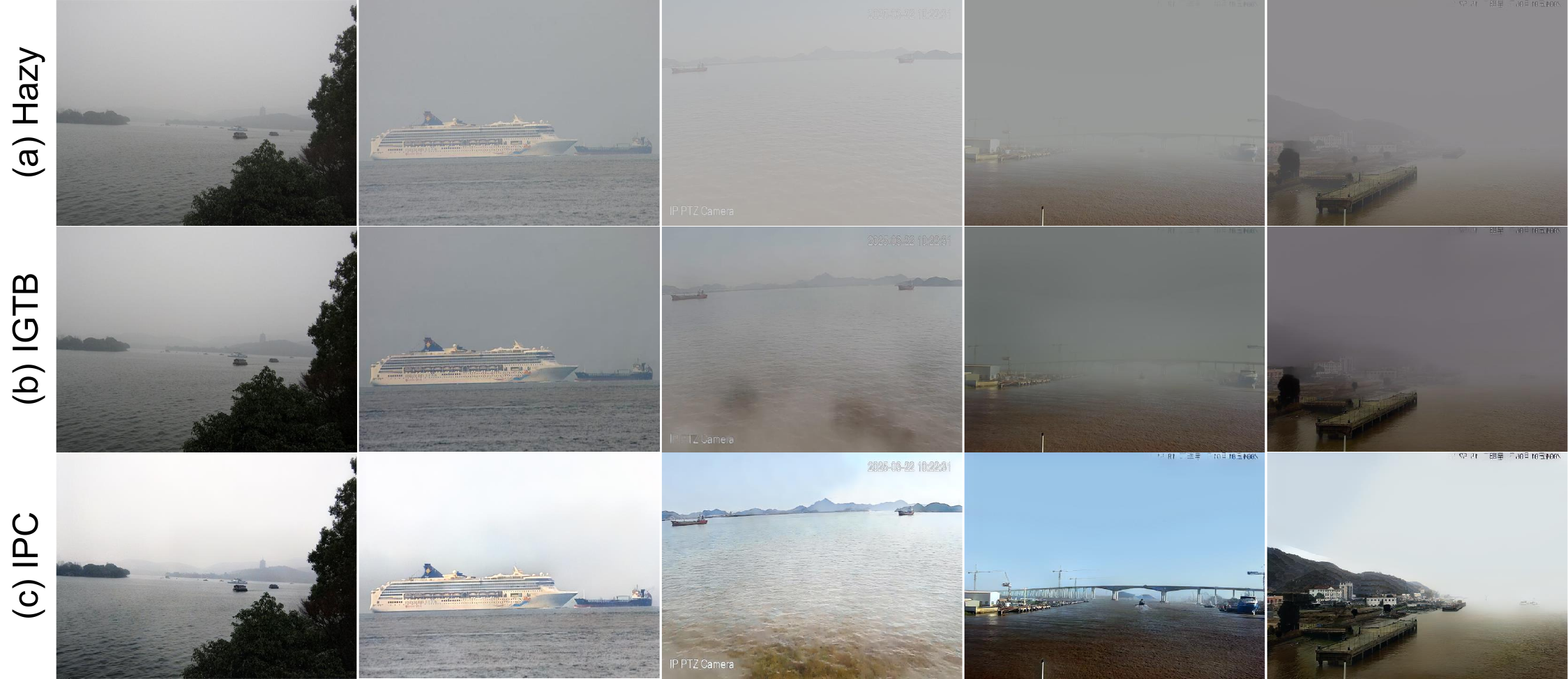}
\caption{Qualitative visual comparison of representative dehazing models on authentic maritime scenes featuring severe visibility degradation. (a) Input scenes with visual degradation. (b) Dehazing results of a lower-scoring model (IGTB), which exhibits substantial residual haze and color darkening. (c) Dehazing results of the top-scoring model (IPC), demonstrating superior visibility restoration, natural contrast, and the effective recovery of distant structural details.}
\label{fig:10}
\end{figure*}

\subsection{Qualitative Cross-Validation on Real Maritime Scenes}
While the synthetic MSVD provides a rigorously controlled environment for establishing and calibrating the proposed visibility metric, validating its sim-to-real generalization capability is essential for practical engineering deployment. In dynamic open-water environments, acquiring perfectly paired hazy images with synchronized meteorological visibility sensor data presents significant objective difficulties. Therefore, to demonstrate the practical relevance and reliability of our metric, we introduce a qualitative cross-validation case study utilizing unannotated real-world maritime hazy images.

Based on the quantitative evaluation results derived from our proposed metric, we deliberately selected two representative dehazing models exhibiting distinct performance levels: the model achieving the highest metric improvement score (IPC) and a model yielding a relatively lower score (IGTB). We subsequently applied these pre-trained models without any domain specific fine tuning to these realistic shipborne scenarios to test their real-world applicability.

As illustrated in Figure \ref{fig:10}, the visual restoration quality aligns with the quantitative predictions of our proposed metric. The top-scoring model (IPC) yields significantly better visual dehazing results on these authentic maritime scenes, successfully recovering distant structural details and restoring natural contrast. Conversely, the lower-scoring model (IGTB) struggles to effectively clear the visual degradation, leaving substantial residual haze and occasionally introducing global color distortion or darkening artifacts. This highly consistent alignment between the quantitative scores and the qualitative visual performance validates that the proposed metric does not merely overfit to specific data distributions. Instead, it reliably evaluates the true structural restoration capabilities of diverse dehazing methods, further demonstrating its feasibility for practical maritime surveillance applications.

\section{Discussions and Future Works}
The primary innovation of this paper lies in the development of a quantitative evaluation framework anchored in object detection performance. To the best of our knowledge, this is the first attempt to provide a specific, quantitative visibility assessment for dehazed maritime images. By establishing preset baseline points and employing piecewise linear interpolation, this framework creates a direct quantitative relationship between the object detection metric mAP50 and physical visibility. This successfully transforms the abstract image quality improvement provided by dehazing methods into a new visibility with practical physical meaning. It not only intuitively demonstrates the methods' performance gains but also significantly enhances the practical value of the evaluation results, which provides critical decision-support data for maritime applications, such as VTS and autonomous navigational safety. Experimental comparative analysis strongly reveals the flaws of traditional metrics in assessing navigational safety value, confirming that the proposed metric effectively avoids the instability and limitations of these traditional metrics. Furthermore, by replacing the benchmark model, the robustness of the quantitative model was verified, successfully demonstrating its superiority in maritime safety scenarios.

Drawing from the key contributions of this study, future research can be advanced in the following directions: 

{\bf{Advanced Simulation and Domain Adaptation: }}The current MSVD simulates uniform haze. Although the MSVD provides precise visibility annotations, there inherently exists a domain gap between the simulated dataset and real-world maritime environments, which are often characterized by dynamic lighting, strong sea surface reflections, and non-uniform haze distributions. To mitigate this gap, future work will focus on integrating complex environmental variables into the rendering engine to further enhance simulation realism. Unsupervised Domain Adaptation (UDA) strategies will be explored to fine-tune models using unannotated real maritime images, thereby enhancing the practical applicability of the proposed visibility quantification metric.

{\bf{Validation with Real-World Maritime Data: }}While the proposed MSVD provides precise visibility labels through simulation, future research will focus on collecting real-world maritime hazy images paired with measured physical visibility ground truths (e.g., using hardware visibility sensors). Pursuing this direction will enable a more comprehensive evaluation of image dehazing models under authentic physical conditions, thereby narrowing the gap between simulation and reality.

{\bf{Integrating Lightweight Design and Low-Cost Deployment: }} Apply the haze removal visibility quantification metric proposed in this paper to evaluate the performance of dehazing models. Future research should focus on assessing and optimizing model parameters, computational complexity, and inference speed, while maintaining the visibility improvement achieved by the proposed metric. Considering the constraints of limited computing capabilities and power constraints of shipborne edge devices in the maritime traffic environment, future work should investigate how to deploy the best-performing and verified models onto shipborne edge computing devices.

{\bf{Directly Empowering Navigation Decision Systems: }}Integrate the calculated visibility directly as an input parameter for maritime traffic management and vessel navigation assistance systems. This will enable the real-time and precise conversion of dehazing method efficacy into practical control decisions, such as automatically triggering changes in fairway control levels based on the variation in new visibility.

\section{Conclusion}

In this study, a quantitative visibility assessment framework, incorporating the proposed MSVD and a novel evaluation metric, is developed to bridge the gap between image restoration and navigational safety in maritime haze environments. Firstly, high-fidelity physical data with precise visibility are acquired through the MSVD, constructed by Unity3D to simulate diverse maritime scenarios. Secondly, to establish a relationship between atmospheric visibility and semantic recognition, a vessel object detection model (YOLO11) is utilized to construct a visibility-detection accuracy correlation table as a calibrated benchmark. Finally, the dehazed visibility is precisely quantified by evaluating model performance against this established benchmark.

The main contribution of this study is the development of a robust, visibility-centric evaluation framework specifically designed for the practical safety requirements of maritime vessel operation. This method effectively reveals the limitations of traditional metrics and provides actionable parameters for nautical risk assessment. A series of extensive experiments demonstrates that the proposed metric maintains effectiveness and robustness across various detection models, and dehazing methods.

Future research will focus on enhancing simulation realism with dynamic environmental factors and optimizing models for lightweight deployment on shipborne edge devices. Furthermore, the framework will be extended to address non-uniform maritime haze. Finally, integrating quantified visibility into navigation assistance systems will enable real-time, precise control decisions for autonomous maritime traffic.

\section*{Acknowledgment}
This research has been supported by the National Natural Science Foundation of China (Nos. 52422111, 52271365), the Excellent Youth Foundation of Hubei Scientific Committee (No. 2024AFA042), and the Key Research and Development Program of Hainan Province (No. ZDYF2026GXJS024).

\section*{CRediT authorship contribution statement}

\textbf{Wentao Feng:} Conceptualization, Methodology, Software, Data curation, Writing – original draft, Visualization; 
\textbf{Guobei Peng:} Methodology, Software, Validation, Data curation; 
\textbf{Wengang Mao:} Writing – review \& editing, Resources, Supervision; 
\textbf{Ryan Wen Liu:} Conceptualization, Writing – review \& editing, Resources, Supervision, Funding acquisition.











\printcredits

\bibliographystyle{cas-model2-names}

\bibliography{main}



\end{document}